\pdfoutput=1

\documentclass[11pt]{article}

\usepackage{acl}

\usepackage{times}
\usepackage{latexsym}

\usepackage[T1]{fontenc}

\usepackage[utf8]{inputenc}

\usepackage{microtype}

\usepackage{inconsolata}

\usepackage{booktabs}
\usepackage{multirow}
\usepackage{graphicx}
\usepackage{makecell}

\title{Your Model Is Not Predicting Depression Well And That Is Why: A Case Study of PRIMATE Dataset}

\author{Kirill Milintsevich$^{1,2}$ \and  Kairit Sirts$^2$ \and Ga\"el Dias$^1$\\
        $^1$Normandie Univ, UNICAEN, ENSICAEN, CNRS, GREYC, France\\
        $^2$Institute of Computer Science, University of Tartu, Estonia\\
        \texttt{\{first\_name\}.\{last\_name\}@\{unicaen.fr$^1$|ut.ee$^2$\}}}

\begin{document}
\maketitle
\begin{abstract}
This paper addresses the quality of annotations in mental health datasets used for NLP-based depression level estimation from social media texts. While previous research relies on social media-based datasets annotated with binary categories, i.e. depressed or non-depressed, recent datasets such as D2S and PRIMATE aim for nuanced annotations using PHQ-9 symptoms. However, most of these datasets rely on crowd workers without the domain knowledge for annotation. Focusing on the PRIMATE dataset, our study reveals concerns regarding annotation validity, particularly for the lack of interest or pleasure symptom. Through reannotation by a mental health professional, we introduce finer labels and textual spans as evidence, identifying a notable number of false positives. Our refined annotations, to be released under a Data Use Agreement, offer a higher-quality test set for anhedonia detection. This study underscores the necessity of addressing annotation quality issues in mental health datasets, advocating for improved methodologies to enhance NLP model reliability in mental health assessments.
\end{abstract}

\section{Introduction}

Applying various NLP techniques to automatically estimate the depression level from social media texts has been a widely researched topic in the field of NLP applied for mental health. Most of these datasets consist of online posts gathered from popular social media platforms, such as Twitter or Reddit. These posts are usually annotated by crowd workers who had only a brief training with a mental health professional (MHP) or sometimes only had access to the annotation instructions. 


While there exist multiple depression-related datasets based on social media texts, most of them only present binary annotation, i.e. whether the user is depressed or not. The most common sources of data are Reddit \citep{losada2016test,yates-etal-2017-depression,pirina-coltekin-2018-identifying} and X (former Twitter) \citep{coppersmith-etal-2014-quantifying,syarif2019study}. Most of the studies use automatic methods of annotations, such as regular expression matching of self-reported terms, like ``I have been diagnosed with depression''. Some of them perform manual verification and annotation either via layman crowd workers~\citep{yates-etal-2017-depression} or by the authors themselves~\citep{coppersmith-etal-2014-quantifying,losada2016test}.

Recently, the interest in more fine-grained depression annotation has emerged. In particular, the two recent datasets D2S~\citep{yadav-etal-2020-identifying} and PRIMATE~\citep{gupta-etal-2022-learning}, identify depressed social media posts from X and Reddit, respectively and annotate them with PHQ-9 symptoms~\citep{kroenke2002phq}. Both datasets have been annotated with the help of crowd workers and later verified by MHPs. However, the verification process was different. For D2S, conflicting annotations were resolved with the majority voting, and the psychiatrist resolved the ties. After that, 100 random samples were selected for quality control and verified by a psychiatrist. Additionally, \citet{zirikly-dredze-2022-explaining} annotated a random sample of D2S with the explanations for each symptom with the help of two MHPs\footnote{\citet{zirikly-dredze-2022-explaining} did not report any conflicts between their annotation and the labels provided with D2S.}, increasing the validity of the data. In the case of PRIMATE, no information is given on the quality control procedure. This raises concerns about the validity of the annotations; thus, we selected PRIMATE for our case study.

In this study, on the example of the PRIMATE dataset, we show that the validity of the annotations for the mental health data is a concern when performed by layman crowd workers. Our MHP reannotated 170 posts from the PRIMATE dataset for the lack of interest or pleasure (anhedonia) symptom. The MHP is the second author of the paper, who is also a practising clinical psychology intern. Our annotations include more fine-grained labels (``mentioned'' vs ``answerable'', as well as an additional ``writer's symptom'' label) as well as spans of texts that serve as evidence of the labels. We observe a high number of false positives in the PRIMATE labels, which can be related to the high difficulty of conceptualizing anhedonia~\citep{rizvi2016assessing}. The annotations are to be released under a Data Use Agreement (DUA), and we believe that it can serve as a higher-quality test set for anhedonia detection.


\section{Dataset}

PRIMATE~\citep{gupta-etal-2022-learning} is a dataset based on the Reddit posts from the r/depression\_help subreddit. Each post is annotated with binary labels for each PHQ-9 question, where ``yes'' means that a post contains the answer to a PHQ-9 question and ``no'' otherwise. The nine symptoms are shortly described as follows: lack of interest or pleasure in doing things (\texttt{LOI}), feeling down or depressed (\texttt{DEP}), sleeping disorder (\texttt{SLE}), lack of energy (\texttt{ENE}), eating disorder (\texttt{EAT}), low self-esteem (\texttt{LSE}), problems with concentrating (\texttt{CON}), hyper or lower activity (\texttt{MOV}), suicidal thoughts (\texttt{SUI}).

The annotation was performed by five crowd workers with additional quality control by an MHP. The information about the annotation procedure or crowd worker training, as well as how exactly the MHPs were involved in the quality control, are not provided in the paper. The only metric on the annotation process is an annotator agreement using Fleiss' kappa, which is reported to be 67\% for initial annotation and 85\% after involvement of the MHPs.

In total, the dataset consists of 2003 posts. Table~\ref{tab:primate_labels} shows the distribution of the labels\footnote{The order of the symptoms in the original work by \citet{gupta-etal-2022-learning} is different from the one of PHQ-9. In our work, we reordered the symptoms to match PHQ-9.}. Note that the exact numbers of labels are slightly different from the ones presented by \citet{gupta-etal-2022-learning}. The dataset is not pre-split into train, validation and test sets; thus, we randomly sample 200 posts for validation and another 200 posts for testing.

\begin{table}[t]
    \centering
    \small
    \begin{tabular}{cll}
        \toprule
         \multirow{2}{*}{\makecell{PHQ-9\\Symptom}} & \multicolumn{2}{c}{Number of Posts} \\
         \cmidrule(lr){2-3}
         & Present & Absent \\
         \midrule
         LOI & 949 & 1054 \\
         DEP & 1664 & 339 \\
         SLE & 374 & 1629 \\
         ENE & 688 & 1315 \\
         EAT & 194 & 1809 \\
         LSE & 1680 & 323 \\
         CON & 195 & 1808 \\
         MOV & 527 & 1476 \\
         SUI & 743 & 1260 \\
         \bottomrule
    \end{tabular}
    \caption{Label distribution in PRIMATE.}
    \label{tab:primate_labels}
\end{table}

\begin{figure}[t]
    \centering
    \includegraphics[width=1\linewidth]{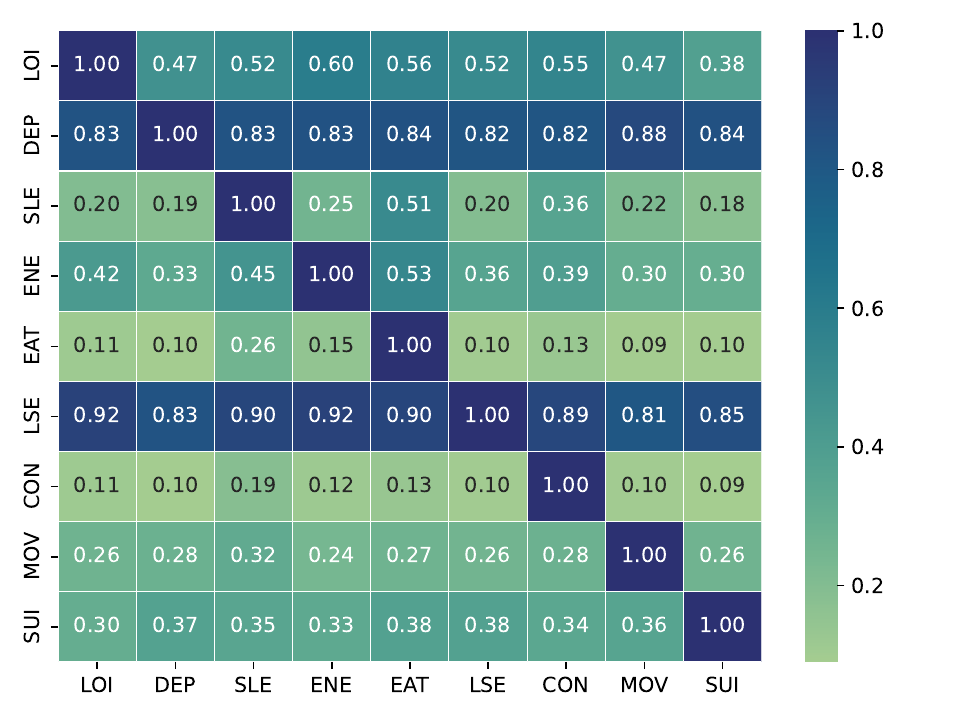}
    \caption{Symptom label co-occurrence matrix of the PRIMATE training set. Each value is normalized column-wise by dividing it by the highest value in the column.}
    \label{fig:cooccurence}
\end{figure}

Figure~\ref{fig:cooccurence} shows the label co-occurrence matrix of the training set. Two symptoms, \texttt{DEP} and \texttt{LSE}, co-occur the most with all the other symptoms, which can be explained by their general prevalence in the dataset. The connection between the lack of interest or pleasure (\texttt{LOI}) and lack of energy (\texttt{ENE}) is also seen in the dataset, which reflects high comorbidity of these symptoms~\citep{van2015association,park2020centrality}.

\begin{table*}[t!]
    \centering
    \begin{tabular}{l|ccccccccc}
        \toprule
        \textbf{Model} & \textbf{LOI} & \textbf{DEP} & \textbf{SLE} & \textbf{ENE} & \textbf{EAT} & \textbf{LSE} & \textbf{CON} & \textbf{MOV} & \textbf{SUI} \\
        \midrule
        DistilBERT & .64 & .88 & .67 & .58 & .60 & .90 & .50 & .67 & .81 \\
        BERT-Base & .55 & .88 & .66 & .55 & .63 & .90 & .46 & .66 & .79 \\
        RoBERTa-Base & .54 & .88 & .70 & .57 & .57 & .90 & .51 & .69 & .85 \\
        RoBERTa-Large & .57 & .86 & .75 & .63 & .65 & .91 & .52 & .71 & .85 \\
        DeBERTa-Base & .58 & .91 & .69 & .52 & .42 & .90 & .36 & .61 & .81 \\
        DeBERTa-Large & .60 & .90 & .68 & .64 & .47 & .91 & .50 & .73 & .83 \\
        \bottomrule
    \end{tabular}
    \caption{Symptom-wise F1-scores on the validation set.}
    \label{tab:main_results}
\end{table*}

\begin{figure*}[t!]
    \centering
    \includegraphics[width=1\linewidth]{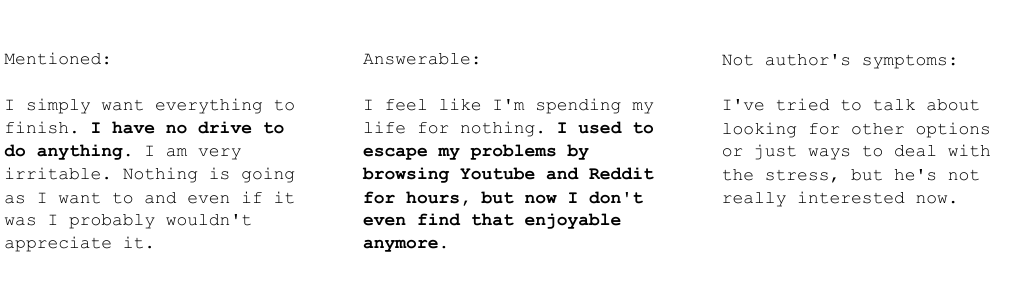}
    \caption{Examples of reannotated posts. Evidences are highlighted in \textbf{bold}.}
    \label{fig:examples}
\end{figure*}

\begin{table*}[t!]
    \centering
    \begin{tabular}{lcccccccccccc}
        \toprule
        \multirow{2}{*}{\textbf{Predictions}} & \multicolumn{4}{c}{Against PRIMATE} & \multicolumn{4}{c}{Against ``mentioned''} & \multicolumn{4}{c}{Against ``answerable''} \\
        \cmidrule(lr){2-5} \cmidrule(lr){6-9} \cmidrule(lr){10-13}
         & A & P & R & F1 & A & P & R & F1 & A & P & R & F1 \\
        \midrule
        DistilBERT & .58 & .56 & .62 & .58 & .56 & .30 & .71 & .42 & .51 & .10 & .75 & .18 \\
        PRIMATE Labels & - & - & - & - & .56 & .27 & .58 & .37 & .54 & .09 & .58 & .15 \\
        \bottomrule
    \end{tabular}
    \caption{Results on the reannotated part of the validation set. Here, \textbf{A} stands for Accuracy, \textbf{P} for Precision, \textbf{R} for Recall, and \textbf{F1} for F1-score for the positive class.}
    \label{tab:annotation}
\end{table*}

\section{Experimental Setup}

In our experiments, we aimed to test how well current pre-trained language models can model the depression symptom detection problem using the PRIMATE dataset. We first chose DistilBERT~\citep{sanh2019distilbert} as a baseline and BERT-Base~\citep{devlin2018bert}, RoBERTa-Base, RoBERTa-Large~\citep{liu2019roberta}, DeBERTa-Base, and DeBERTa-Large~\citep{he2020deberta} as higher-performing models. In particular, DeBERTa has shown constant improvements in various NLP tasks and replaced BERT and RoBERTa as the state-of-the-art model for many of them\footnote{\url{https://gluebenchmark.com/leaderboard}}. 

For fine-tuning, we used the implementation from Transformers library~\citep{wolf-etal-2020-transformers}. Each model consists of a pre-trained encoder with a classification head on the top of the \texttt{[CLS]} token. The classification head is represented by a linear layer; in the case of DeBERTa, another linear layer followed by GELU~\citep{hendrycks2016gaussian} is added before the classification head. We trained each model for 20 epochs using AdamW optimizer with the learning rate of $2e^{-5}$, $\epsilon$ of $1e^{-6}$, $\beta_1, \beta_2$ of $(0.9, 0.999)$, and weight decay $\lambda$ of $0.01$. Additionally, a linear learning rate scheduler is applied with a warmup ratio of $0.1$. Finally, the training batch size was set to $16$.


\section{Results and Discussion}

Table~\ref{tab:main_results} shows that larger models, such as RoBERTa-Large and DeBERTa-Large, perform better for \texttt{ENE}, \texttt{LSE}, \texttt{MOV}, and \texttt{SUI}. Additionally, \texttt{DEP} shows slight improvement with DeBERTa models, however, decreased performance for \texttt{EAT}. RoBERTa models perform better for \texttt{SLE} and \texttt{SUI} prediction. Nevertheless, DistilBERT sets a strong baseline and performs on par with larger models overall. Finally, \texttt{LOI} shows a decrease in performance for all the models compared to the DistilBERT.

We investigate the diminished performance of the \texttt{LOI} symptom since it is a core symptom of a major depressive disorder~\citep{dsm2013} and shows unstable results for our models. Furthermore, \texttt{LOI} is one of the symptoms of schizophrenia~\citep{dsm2013} and is associated with both anxiety and depression~\citep{winer2017mapping}. Thus, we selected a subset of 170 posts from the validation set based on the DistilBERT predictions: if at least one symptom was predicted incorrectly, the post was selected. Next, an MHP read all the posts in the subset and labelled them for the presence of loss of interest or pleasure (\texttt{LOI}). The MHP assigned three labels to each post: a) ``mentioned'' if the symptom is talked about in the text, but it is not possible to infer its duration or intensity; b) ``answerable'' if there is clear evidence of anhedonia; c) ``writer's symptoms'' which shows whether the author of the post discusses themselves or a third person. Additionally, the MHP selected the part of the text that supports the positive label.

Figure~\ref{fig:examples} shows examples for the reannotated posts\footnote{All example posts are paraphrased for privacy.}. The first example is labelled as ``mentioned'' since it contains evidence of a symptom but does not contain information about the \emph{loss} of interest. The second example is labelled as ``answerable'' because it is possible to infer that the person used to have interest in what they were doing before but lost it at some point in time. Finally, the last example shows the post without signs of \texttt{LOI} that describes the condition of another person.

Table~\ref{tab:annotation} shows accuracy, precision, recall and F1-score for positive class against different sets of labels on our manually reannotated subset. DistilBERT, when measured against ``mentioned'' and ``answerable'' labels, performs considerably worse than against original labels from PRIMATE. It is unsurprising given the extremely low agreement between these sets of labels with Cohen's kappa of 9\% and 3\%, respectively. Furthermore, the most common error type is a false positive, i.e., a symptom marked as present in PRIMATE when our MHP found no evidence of it in the text. Additionally, using PRIMATE labels as predictions and comparing their performance against our labels shows lower performance than the DistilBERT model.

Considering the ``writer's symptom'' label, in 18 out of 170 selected posts, the author describes a symptom of another person rather than themselves. This raises the question of how these posts should be annotated and whether they should be included in the dataset at all. We suspect that the language of describing one's condition or feelings in the first person is different from the third person. We leave this question for future debate and assign ``mentioned'' and ``answerable'' labels to the posts describing a third person in the same manner as to the personal posts.

Our findings are consistent with the original results presented by \citet{gupta-etal-2022-learning}. Similar to our experiment, they also trained a classifier based on the BERT-Base model and reported low MCC for \texttt{LOI}. However, we provided the evidence that this might be caused by annotation errors. Additionally, we noticed that many posts that were mistakenly labelled with \texttt{LOI} are more closely related to the ``inner tension'' symptom from the Montgomery-Åsberg Depression Rating Scale (MADRS)~\citep{montgomery1979new}.

While we agree that our reannotated test set is also, to some extent, susceptible to errors, we believe that it serves as a more reliable benchmark for the anhedonia symptom. A more fine-grained, evidence-based labelling scheme reduces the risk of mislabelling and is more transparent for further verification. Finally, it lays the foundation for future collaboration to produce a higher-quality Reddit-based dataset for depression symptom estimation.

\section{Conclusion}

In conclusion, this study highlights the importance of evaluating and enhancing the quality of annotations in mental health datasets, particularly within the context of automated depression level estimation from social media texts. While recent datasets such as PRIMATE introduce commendable efforts toward nuanced annotations using PHQ-9 symptoms, our examination of the PRIMATE dataset reveals concerns about annotation validity, specifically regarding the lack of interest or pleasure symptom. Through careful reannotation by a mental health professional, we discerned a considerable number of false positives among the original labels indicative of challenges in conceptualizing anhedonia.

The findings presented here advocate for a more rigorous and standardized approach to mental health dataset annotation, emphasizing the need for greater involvement of domain experts in the annotation process. The release of our refined annotations under a Data Use Agreement (DUA) contributes a valuable resource for future research, offering a higher quality test set for anhedonia detection. Moving forward, a concerted effort toward refining annotation methodologies and promoting collaboration between domain experts and NLP practitioners is imperative to foster advancements in this crucial intersection of technology and mental health research.

\section{Availability of Data}

The instructions for accessing the annotations presented in this paper can be found here: \url{https://github.com/501Good/primate-anhedonia}.

\section{Ethical Considerations}

According to \citet{benton-etal-2017-ethical}, studies involving user-generated content are exempt from Institutional Review Board (IRB) requirements if the data source is public and user identities are not identifiable. We access and use the data according to the Data Use Agreement provided with the PRIMATE dataset. Finally, we are going to release our annotations under another Data Use Agreement and separate them from the original PRIMATE data. We also acknowledge that no automatic system can replace a real mental health professional and cannot be used as a sole instrument of diagnostics.

\section{Limitations}

We acknowledge the limitations inherent in our work and findings. First, the manually annotated explanations serve as a proxy for what clinicians might find informative in assessing Reddit posts flagged as depressive. While evaluating the informativeness of explanations in a true clinical setting would provide more insight, it falls beyond the scope of this paper. Furthermore, our reannotation was carried out by only one mental health professional, which does not allow for performing an inter-annotator agreement analysis. However, we believe that our evidence-based labelling scheme partially mitigates this problem. Finally, anhedonia is extremely challenging to conceptualize and binary labels may not be the best choice in situations when the difference between the presence or absence of the symptom is marginal. In this case, labels based on the Likert scale, as in PHQ-9, would be more appropriate and allow us to capture the intensity of the symptom more accurately. Furthermore, different demographics, for example, adolescents and adults, express signs of anhedonia differently~\citep{watson2020understanding}.

\section*{Acknowledgements}
This research was supported by the Estonian Research Council Grant PSG721 and the FHU A$^2$M$^2$P project funded by the G4 University Hospitals of Amiens, Caen, Lille and Rouen (France). The calculations for model’s training and inference were carried out in the High Performance Computing Center of the University of Tartu~\citep{https://doi.org/10.23673/ph6n-0144}.

\bibliography{anthology,custom}

\appendix

\end{document}